\documentclass{llncs}

\usepackage[T1]{fontenc}
\usepackage{graphicx}

\usepackage{url}
\usepackage{listings}
\usepackage{xspace}
\usepackage{svg}
\usepackage{hyperref}
\usepackage{xcolor}
\usepackage{multirow}
\usepackage{colortbl}
\usepackage{subcaption}
\usepackage{enumitem}
\usepackage{amsmath}
\usepackage{amsfonts}
\usepackage{amssymb}
\usepackage[ruled,linesnumbered ]{algorithm2e}
\usepackage{booktabs}

\usepackage[para]{footmisc}

\SetKwComment{Comment}{/* }{ */}

\definecolor{chromeyellow}{rgb}{1.0, 0.65, 0.0}

\definecolor{darkgreen}{rgb}{0.0, 0.5, 0.0}

\newcommand{\owlclass}{\small {\textit{owl:Class}}\xspace}
\newcommand{\owlclasses}{\small {\textit{owl:Classes}}\xspace}

\newcommand{\dspipeline}{\small {\textit{ds:Pipeline}}\xspace}
\newcommand{\dstask}{\small {\textit{ds:Task}}\xspace}
\newcommand{\dsatomictask}{\small {\textit{ds:AtomicTask}}\xspace}
\newcommand{\dsmethod}{\small {\textit{ds:Method}}\xspace}
\newcommand{\dsatomicmethod}{\small {\textit{ds:AtomicMethod}}\xspace}

\newcommand{\dsdataentity}{\small {\textit{ds:DataEntity}}\xspace}
\newcommand{\dsdatastructure}{\small {\textit{ds:DataStructure}}\xspace}
\newcommand{\dssinglevalue}{\small {\textit{ds:SingleValue}}\xspace}

\newcommand{\dshasnexttask}{\small {\textit{ds:hasNextTask}}\xspace}
\newcommand{\dshasinputdatapath}{\small {\textit{ds:hasInputDataPath}}\xspace}
\newcommand{\rdfssubclassof}{\small {\textit{rdfs:subClassOf}}\xspace}
\newcommand{\owlindiv}{\small {\textit{owl:Individual}}\xspace}
\newcommand{\owlindivs}{\small {\textit{owl:Individuals}}\xspace}
\newcommand{\trainmethod}{\small {\textit{ml:TrainMethod}}\xspace}
\newcommand{\traintask}{\small {\textit{ml:Train}}\xspace}
\newcommand{\plottask}{\small {\textit{visu:PlotTask}}\xspace}
\newcommand{\canvastask}{\small {\textit{visu:CanvasTask}}\xspace}

\newcommand{\rdflib}{\texttt{rdflib}\xspace}
\newcommand{\pyshacl}{\texttt{pySHACL}\xspace}

\newcommand{\footnotehref}[1]{\footnote{\href{#1}{#1}}}
\newcommand{\refpar}[2]{(Paragraph \hyperref[#1]{\textit{#2}})}

\newcommand{\exekglib}[0]{\textit{ExeKGLib}\xspace}

\newcommand{\eg}[0]{\textit{e.g.},\xspace}
\newcommand{\ie}[0]{\textit{i.e.},\xspace}

\newcolumntype{P}[1]{>{\centering\arraybackslash}p{#1}}
\newcolumntype{M}[1]{>{\centering\arraybackslash}m{#1}}

\newcommand\BibTeX{{\rmfamily B\kern-.05em \textsc{i\kern-.025em b}\kern-.08em
T\kern-.1667em\lower.7ex\hbox{E}\kern-.125emX}}

\begin{document}
\title{ExeKGLib: A Platform for Machine Learning Analytics based on Knowledge Graphs}
\titlerunning{ExeKGLib: A Platform for ML Analytics based on KGs}
% If the paper title is too long for the running head, you can set
% an abbreviated paper title here
%

\author{Antonis Klironomos\inst{1,2} \and
Baifan Zhou\inst{4,3} \and
Zhipeng Tan\inst{1,5} \and
Zhuoxun Zheng\inst{1,3} \and
Mohamed H. Gad-Elrab\inst{1} \and
Heiko Paulheim\inst{2} \and
Evgeny Kharlamov\inst{1,3}}
\authorrunning{A. Klironomos et al.}
% First names are abbreviated in the running head.
% If there are more than two authors, 'et al.' is used.
%
\institute{Bosch Center for AI, Germany\\
% \email{\{antonis.klironomos,zhuoxun.zheng,mohamed.gad-elrab,evgeny.kharlamov\}@de.bosch.com} \and
\email{firstName.lastName@de.bosch.com} \and
University of Mannheim, Germany\\
\email{heiko@informatik.uni-mannheim.de} \and
University of Oslo, Norway\\
\email{baifanz@ifi.uio.no} \and
Oslo Metropolitan University, Norway \and
RWTH Aachen, Germany\\
\email{Zhipeng.tan1@outlook.com}}

% \author{Antonis Klironomos$^{1,2}$,
% Baifan Zhou$^{4,3}$,
% Zhipeng Tan$^{1,5}$, 
% Zhuoxun Zheng$^{1,3}$,
% Gad-Elrab Mohamed$^{1}$
% Heiko Paulheim$^{2}$,
% Evgeny Kharlamov$^{1,3}$}
% %\\[2ex]
% %\small{$^{1}$Bosch Center for Artificial Intelligence, Germany,
% %$^{2}$University of Mannheim, Germany,
% %$^{3}$University of Oslo, Norway,
% %$^{4}$Oslo Metropolitan University, Norway,
% %$^{5}$RWTH Aachen, Germany}}
% \institute{Bosch Center for AI, Germany,
% $^{2}$University of Mannheim, Germany,
% $^{3}$University of Oslo, Norway,
% $^{4}$Oslo Metropolitan University, Norway,
% $^{5}$RWTH Aachen, Germany}
%
\authorrunning{A. Klironomos et al.}
% First names are abbreviated in the running head.
% If there are more than two authors, 'et al.' is used.

% % First names are abbreviated in the running head.
% % If there are more than two authors, 'et al.' is used.
% %
% \institute{Princeton University, Princeton NJ 08544, USA \and
% Springer Heidelberg, Tiergartenstr. 17, 69121 Heidelberg, Germany
% \email{lncs@springer.com}\\
% \url{http://www.springer.com/gp/computer-science/lncs} \and
% ABC Institute, Rupert-Karls-University Heidelberg, Heidelberg, Germany\\
% \email{\{abc,lncs\}@uni-heidelberg.de}}
%
\maketitle              % typeset the header of the contribution
\begin{abstract}
Nowadays machine learning (ML) practitioners have access to numerous ML libraries available online. Such libraries can be used to create ML pipelines that consist of a series of steps where each step may invoke up to several ML libraries that are used for various data-driven analytical tasks. Development of high-quality ML pipelines is non-trivial; it requires training, ML expertise, and careful development of each step. At the same time, domain experts in science and engineering may not possess such ML expertise and training while they are in pressing need of ML-based analytics. In this paper, we present our \exekglib, a Python library enhanced with a graphical interface layer that allows users with
% coding skills \antonis{remove coding skills?} and
minimal ML knowledge to build ML pipelines. This is achieved by relying on knowledge graphs that encode ML knowledge in simple terms accessible to non-ML experts. \exekglib also allows improving the transparency and reusability of the built ML workflows and ensures that they are executable. We show the usability and usefulness of \exekglib by presenting real use cases.

% Many machine learning (ML) libraries are accessible online for ML practitioners. Typical ML pipelines consist of a series of steps where each step invokes several ML libraries. The production of high-quality ML pipelines requires training, ML expertise, and careful development of each step. At the same time, many domain experts in science and engineering are learning ML and want to apply ML to answer their questions. However, there has been a limited number of ML libraries for experts in non-ML disciplines. In this paper, we present \ExeKGLib, a Python library that allows users with coding skills and minimal ML knowledge to build ML pipelines. \ExeKGLib relies on knowledge graphs to improve the transparency and reusability of the built ML workflows, and to ensure that they are executable. We show the usability and usefulness of \ExeKGLib by conducting a user study and by presenting real use cases.

%\keywords{Machine learning \and Knowledge graphs \and Python library}
\end{abstract}
\section{Introduction}
\label{sec:intro}

% ML is becoming popular
Thanks to the remarkable progress in Computer Science and specifically the field of machine learning (ML), there is a plethora of ML algorithms and corresponding libraries publicly accessible. Both in academia and industry, the popularity of ML is continuously increasing \cite{sarkerMachineLearningAlgorithms2021}. Many experts in other domains are also learning ML and want to apply it to solve their specific problems, \eg biologists \cite{libbrechtMachineLearningApplications2015,kimInfectiousDiseaseOutbreak2021}, oncologists \cite{kourouMachineLearningApplications2015,abreuPredictingBreastCancer2016}, and engineers \cite{mengMachineLearningAdditive2020,rangel-martinezMachineLearningSustainable2021}.

% there is a need to allow non-ML experts to do ML
The development and optimization of ML workflows can be complex and time-consuming. Mastering sophisticated ML skills without prior knowledge requires a considerable amount of time.
This poses a barrier for domain experts who want to use ML but have limited ML background and limited time for learning ML skills. 
Thus, there is a need in the community and particularly in the industry to lower the barrier of non-ML experts using ML, by providing a user-friendly way that does not require excessive training %in ML 
for ML pipeline development. 
This is especially pressing in data-intensive smart manufacturing where rapid and easy development of high-quality ML pipelines is vital for scaling optimization solutions for production and products by engaging large numbers of production experts in the ML pipeline development. 
% finding a user-friendly way to allow experts with limited training in ML to use it.

% previous libraries did not solve the problem
The idea of simplifying ML pipeline development is not new and there have been multiple attempts and solutions developed so far. The most prominent group of solutions is AutoML~\cite{hutterAutomatedMachineLearning2019} that, despite a large success, offers only limited customizability for modifications of ML pipelines and other important tasks are largely ignored such as customized data visualization, data preprocessing, statistical methods, feature engineering, etc., which is a limiting factor, especially in smart manufacturing.
Most existing free tools, such as RapidMiner~\cite{hofmann2016rapidminer} and KNIME~\cite{BCDG+07}, do not incorporate linked open data (LOD) in their functionality. Although RapidMiner includes semantic annotations for ML pipelines, it does not leverage LOD as extensively as our tool, which utilizes LOD throughout the entire pipeline lifecycle. Tool characteristics are compared in Table~\ref{tab:tool-feature-comparison}.
% \baifan{do we need a table ti summarise the different tools and novelty of ExeKGLib? and justify why it is necessary?}
% \antonis{added table on related work section and referring to it here}

% Other tools such as (a) Weka, (b) Amazon Sage Maker, or (c) Google AutoML provide a graphic user interface (GUI) but have limited transparency for domain experts and either they do not provide open-source code libraries (in case of (b) and (c)) or they are developed in programming languages that are not the most popular for ML (in case of (a)).

% \antonis{TODO: modify above paragraph according to the changed related work section}

%Additionally, they do not rely on Knowledge Graphs (KGs) which help with comprehending and reusing the ML pipelines.

%Existing similar tools such as the ones described in the official AutoML book~\cite{hutterAutomatedMachineLearning2019}, have similar goals of lowering the barrier of using ML, but they offer limited customizability for modifications of ML pipelines and other important tasks are largely ignored such as customized data visualization, data preprocessing, statistical methods, feature engineering, etc. Other tools such as Amazon Sage Maker\footnote{https://aws.amazon.com/sagemaker} or Google AutoML\footnote{https://cloud.google.com/automl} provide a graphic user interface (GUI) but do not provide open-source code libraries. Additionally, they do not rely on Knowledge Graphs (KGs) which help with comprehending and reusing the ML pipelines.

\begin{figure*}[t]
    % \vspace{-4ex}    
    \centering
    % \makebox[\textwidth][c]{
        \includegraphics[width=1\textwidth]{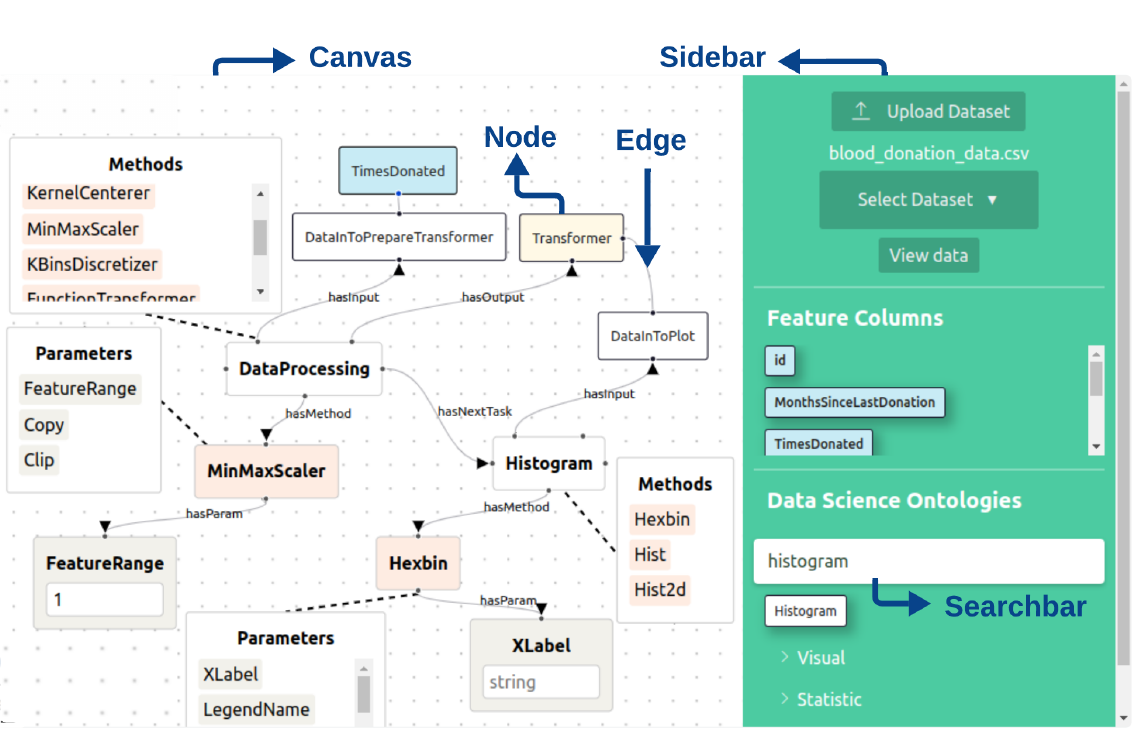}
      % }
    % \vspace{-6ex}
    \caption{\exekglib's Graphical User Interface (GUI)
    % \baifan{the resolution is bit low, difficult to see the text}
    % \baifan{I feel the paper is now really strong with both the GUi and use cases :)}
    % \antonis{changed the figure}
    }
    \label{fig:gui-elements}
    % \vspace{-6ex}    
\end{figure*}

% we propose our tool
To address these challenges, we propose \exekglib, a Python library that lowers the barrier to using ML, allowing people with
% coding skills and 
minimal ML expertise to exploit the potential of ML through a user-friendly graphical interface. We call the library \exekglib because it relies on the construction of Knowledge Graphs (KGs) that can be translated to executable data pipelines~\cite{zhengExecutableKnowledgeGraphs2022a}. The KGs provide a formalized description of the ML pipelines, improving transparency and reusability. In addition, these KGs are based on a schema, and their components are connected to SHACL constraints. This ensures that the ML pipelines are executable. \exekglib works in two steps: (1) generating executable ML pipelines via KGs, (2) converting them to Python scripts and executing the scripts. \exekglib supports a variety of methods for data visualization, data preprocessing and feature engineering, and ML modeling. It also has an interface of extension to allow users to extend to other libraries and customized scripts.

% \antonis{TODO: Add brief mention to the existing GUI and add a screenshot}
\exekglib has been successfully evaluated in an industrial setting at Bosch and showed its great potential. 
The tool is combined with an internal GUI to faciliate its usage as %which we plan to publish soon.
% A version of the GUI is 
shown in Fig.~\ref{fig:gui-elements}~\cite{nocodeml}.
With this paper, we aim to share \exekglib with the community as a valuable resource~\cite{zhengExecutableKnowledgeGraphs2022a}. In particular, \exekglib is the backbone of our semantic ML solution, SemML~\cite{DBLP:journals/jim/ZhouPRKM22,DBLP:conf/cikm/ZhengZZSK22a,DBLP:conf/cikm/ZhouZZSSKK22,zhouSemMLFacilitatingDevelopment2021}, that has been used, \eg for welding quality monitoring and optimization for resistance spot welding and hot staking, and plastic data analytics, and now extensively evaluated in several EU publicly funded projects and Bosch internal projects. 
% \heiko{It would be good to add a few sentences about which actual benefits the applications at BOSCH have and/or which actual problems are solved.} 
% \antonis{added in the next sentences}
For instance, the welding use case has to do with the automated welding of car bodies in assembly lines. Even one low-quality welding spot can cause the stop of the whole assembly line. Traditional methods for monitoring welding quality involve destroying welded car bodies which is extremely expensive. Also, the estimation of welding quality requires working hours and expertise from multiple practitioners. By introducing semantically annotated ML-based methods to predict welding quality, Bosch aims to minimize the requirement for destroyed car bodies and the time consumed by experts. This leads to decreased waste and promotes more cost-effective and sustainable manufacturing practices.
The usage of \exekglib at Bosch has facilitated the creation and modification of ML workflows by experts from various fields such as welding, sensor engineering, and ML. Furthermore, due to its semantic aspect, \exekglib has increased the quality of communication between experts.

% \heiko{add another paragraph enumerating the technical aspects of the contribution: a schema, a library for translating RDF to executable code, etc.}
% \antonis{added below}
Our contributions are summarized below: %\gad{rewritten}
\begin{itemize}[topsep=3pt,parsep=0pt,partopsep=0pt,itemsep=0pt,leftmargin=*]
    \item We provide one upper-level and three semi-automatically generated lower-level KG schemata that describe ML pipelines.
    \item We provide a SHACL shapes graph for the validation of ML pipelines.
    \item As part of \exekglib's functionality:
    \begin{itemize}[topsep=3pt,parsep=0pt,partopsep=0pt,itemsep=0pt,leftmargin=*]
        \item We use the provided KG schemata and SHACL shapes graph to create and validate executable KGs (ExeKGs).
        \item We automate the process of translating pipelines from RDF to Python code.
        \item We offer an LLM-enhanced graphical user interface (GUI), a simple coding interface, and a command line interface (CLI).
    \end{itemize}
\end{itemize}

%\baifan{(referring to the 2022ESWC resource of Python lib paper) here we need to say where the library is used. you can say it is developed from and used in several projects, such as RSW monitoring, HS monitoring,  (1) at Bosch, backbone for SemML, (2) coding tool for engineers who want to learn ML, (3) part of OntoCommons  }
%\antonis{This library is used (1) as a core component of a larger framework (SemML)~\cite{zhouSemMLFacilitatingDevelopment2021} at Bosch for improving the welding quality in the automotive industry, (2) in the OntoCommons EU project, and (3) as a means for engineers to learn ML.}

% \exekglib can be used by a wide range of users and in a variety of scenarios, for which we give several examples (Section \ref{sec:impact-and-usage}): (1) Domain experts with programming skills and basic knowledge of ML that want to do ML, such as scientists and industrial practitioners;
% (2) Software applications incorporating ML to boost their functionality;
% (3) ML experts that need to manage a large amount of ML pipelines who need an automated way of generating ML pipelines;
% (4) ML experts in interdisciplinary projects who need to explain ML pipelines to domain experts;
% (5) Teachers and students for teaching and learning ML.

% outline of the paper
The paper is organized as follows: 
Section~\ref{sec:related-work} reviews related works with a similar goal;
Section~\ref{sec:exekglib} elaborates \exekglib's functionality, design, architecture, implementation, and GUI;
% Section~\ref{sec:gui} describes the GUI's design, features, and LLM-based enhancements;
Section~\ref{sec:evaluation} discusses the usage of \exekglib with real use cases, and stresses the proposed tool's impact;
Section~\ref{sec:conclusion} concludes the paper.

% \begin{figure*}[t]
% % \vspace{-2ex}
%     \includegraphics[width=1\textwidth]{content/figures/NewUISample2.png}
%     % \vspace{-6ex}
%     \caption{Example of the pipeline creation flow through \exekglib's GUI}
%     \label{fig:ui}
%     % \vspace{-5ex}
% \end{figure*}

\section{Related work}
\label{sec:related-work}
% \vspace{-4ex}

Semantic technologies are increasingly used to interpret complex software and ML models, for instance, by generating KGs for ML execution~\cite{draschnerDistRDF2MLScalableDistributed2021} or code comprehension~\cite{khanKnowledgeGraphGeneration2021a}. Literature reviews~\cite{breit2023combining,ristoski2016semantic,seeligerSemanticWebTechnologies2019a,tiddiKnowledgeGraphsTools2022} affirm the growing role of semantics in ML, while also highlighting challenges such as the manual creation of KGs and the need for more integrated knowledge representation across domains~\cite{tiddiKnowledgeGraphsTools2022}. Against this backdrop, this section delves into existing ontologies designed for describing ML experiments and code artifacts, and subsequently examines tools and libraries that aid in constructing ML workflows.

\medskip
\noindent \textbf{Similar Ontologies.}
While several ontologies address the description of ML experiments and artifacts, they exhibit limitations when viewed against the requirements for our KG schemata. For instance, ontologies for ML experiments such as Exposé~\cite{vanschorenExperimentDatabases2012}, ML-Schema~\cite{publioMLSchemaExposingSemantics2018}, MEX Vocabulary~\cite{estevesMEXVocabularyLightweight2015}, and PROV-ML~\cite{souzaProvenanceDataMachine2019} primarily emphasize experiment tracking (with some incorporating provenance) rather than explicit pipeline representation, and can introduce metadata overhead. In the context of code, the Software Ontology (SWO)~\cite{maloneSoftwareOntologySWO2014} inadequately captures ML pipeline data flow, while the Semanticscience Integrated Ontology (SIO)~\cite{dumontierSemanticscienceIntegratedOntology2014}, though also applied to code~\cite{abdelazizToolkitGeneratingCode2021}, is too low-level, lacking specific data science concepts or an ML task hierarchy. In data mining, OntoDM~\cite{panovOntoDMOntologyData2008} overlooks many ML tasks and workflow details. Similarly, the Machine Learning Schema Ontology (MLSO) semantically represents datasets and associated ML pipelines to generate MLSeaKG, a KG of ML datasets and pipelines~\cite{dasoulas2024mlsea,dasoulas2024mlseascape}. However, it focuses primarily on representing datasets and lacks a granular representation of the ML pipelines. Furthermore, this approach is geared towards creating semantic layers for ML metadata discovery, not generating ML pipelines as KGs throughout their entire lifecycle of creation, validation, and executability, as our \exekglib framework does. Even DMOP~\cite{keetDataMiningOPtimization2015}, which is arguably the most similar to our approach in aiming to optimize data mining processes by defining elements like data sources and algorithms, tends to focus on technical algorithm characteristics instead of the high-level representation needed. These limitations highlight the need for a schema capable of effectively modeling complex ML tasks and workflows with suitable detail and high-level concepts, avoiding unnecessary overhead and ensuring essential pipeline representations are not missing.

\medskip
\noindent \textbf{Similar Tools or Libraries.}
While AutoML~\cite{hutterAutomatedMachineLearning2019} and Declarative ML~\cite{molinoDeclarativeMachineLearning2021} offer methods for ML workflow construction, existing open-source tools like Weka~\cite{frank2005weka}, RapidMiner~\cite{mierswaYALERapidPrototyping2006}, Orange~\cite{JMLR:demsar13a}, KNIME~\cite{BCDG+07}, and Ludwig~\cite{Molino2019} typically utilize GUIs or YAML interfaces but may present limitations such as Java-centric designs or narrower scopes (\eg Ludwig's deep learning focus). Notably, though RapidMiner incorporated semantic annotation,~\cite{kietzSemanticsLetNot2014a} these tools do not generate executable KGs. In stark contrast, our tool, \exekglib, uniquely represents ML pipelines using LOD formats, specifically RDF and OWL, rather than proprietary or plain serialization methods. This allows \exekglib to employ schemata for guided pipeline creation, enhance pipeline understanding and reusability (see Section~\ref{sec:use-cases}), and improve overall management (\eg through repository integration and batch creation of pipelines as KGs for rapid experimentation via its coding interface). Consequently, \exekglib stands as the first free, open-source library empowering programmers, including those with limited ML expertise, to create, validate, and execute custom, extensible ML pipelines as KGs. This offers significant benefits such as open pipeline formats, increased transparency and reusability, and streamlined, quick experimentation (see Table~\ref{tab:tool-feature-comparison}).

\begin{table*}[t]
    \centering
    \caption{Comparing \exekglib features with similar free open-source tools. Unique features of \exekglib are denoted by `\( \blacktriangle \)'.}
    \label{tab:tool-feature-comparison}
    \begin{tabular}{l c c c c c c} % Changed column specification
        \toprule % Use booktabs for better spacing
        \textbf{Feature} & 
        \rotatebox{45}{Weka} & 
        \rotatebox{45}{RapidMiner} & 
        \rotatebox{45}{Orange} & 
        \rotatebox{45}{KNIME} & 
        \rotatebox{45}{Ludwig} & 
        \rotatebox{45}{\textbf{\exekglib}} \\ % Moved exekglib header to the end
        \midrule % Use booktabs rule
        Python Implementation                     &           &                 & $\checkmark$ &                 & $\checkmark$ & $\checkmark$ \\
        Semantic Validation              &           & $\checkmark$    &                 &                 &                 & $\checkmark$ \\
        Semantic Creation \& Execution \( \blacktriangle \) &           &                 &                 &                 &                 & $\checkmark$ \\
        Batch Generation \& Execution \( \blacktriangle \)  &           &                 &                 &                 &                 & $\checkmark$ \\
        Open Pipeline Format                      &           & $\checkmark$    &                 &                 & $\checkmark$ & $\checkmark$ \\
        LOD Pipeline Format \( \blacktriangle \)       &           &                 &                 &                 &                 & $\checkmark$ \\
        % \midrule % Use booktabs rule
        % Statistical ML                            & $\checkmark$ & $\checkmark$    & $\checkmark$ & $\checkmark$    &                 & $\checkmark$ \\
        % Non-parametric ML                         & $\checkmark$ & $\checkmark$    & $\checkmark$ & $\checkmark$    &                 & $\checkmark$ \\
        % Neural Network ML                         & $\checkmark$ & $\checkmark$    & $\checkmark$ & $\checkmark$    & $\checkmark$ & $\checkmark$ \\
        \bottomrule % Use booktabs rule
    \end{tabular}
    % \vspace{-4ex}
\end{table*}
% \input{content/3-decisions-and-func.tex}
% \input{content/3-functionality.tex}
% \section{The \exekglib}
\section{Executable Knowledge Graphs Library}
\label{sec:exekglib}

In this section, we present the functionality and software architecture of the proposed Python library.
% \footnotehref{https://github.com/boschresearch/ExeKGLib}
\exekglib relies on KG schemata to construct Exe\-KGs (which represent ML pipelines) and execute them. It also utilizes \pyshacl to validate the executability of the constructed KGs. The aforementioned processes use the \rdflib Python library combined with SPARQL queries to find and create KG components. These components are automatically mapped to and stored as Python objects based on a custom class hierarchy that corresponds to the \owlclass hierarchy defined in the KG schemata.

\subsection{Functionality}
\label{sec:functionality}

\begin{figure*}[t]
% \vspace{-4ex}
    \centering
    \includegraphics[width=1\textwidth]{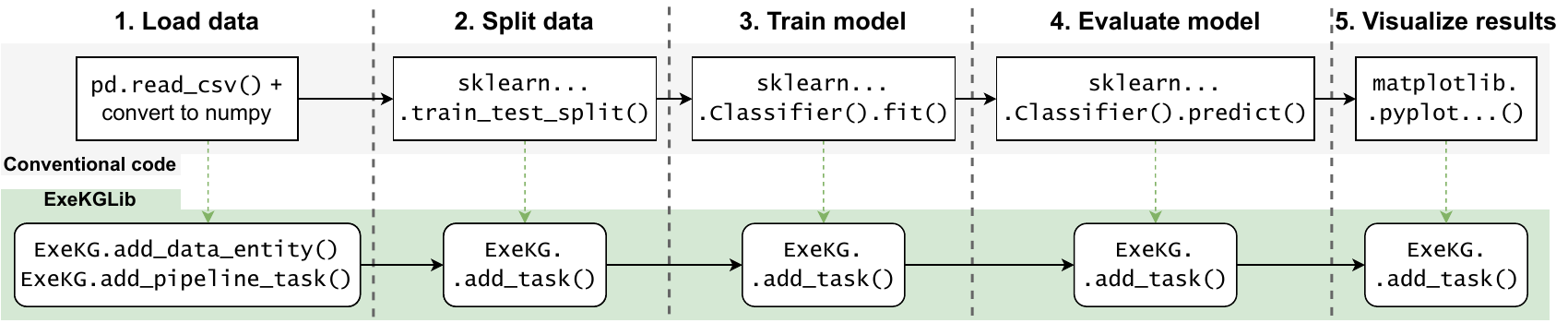}
    % \vspace{-5ex}
    \caption{Comparison between conventional code and \exekglib for a classification problem --- \exekglib can be learned faster. Fig.~\ref{fig:kg-construction} details how this is achieved.}
    \label{fig:code-comparison}
    % \vspace{-2ex}
\end{figure*}

\exekglib's functionality and practicality are illustrated in Fig.~\ref{fig:code-comparison}, which compares the design of a generic classification pipeline in the conventional setup versus using \exekglib.
% \exekglib's functionality can be communicated by showing the tool's practical advantages on a generic classification task (Fig.~\ref{fig:code-comparison}).
% The upper part of the figure shows 
In the conventional setup, shown in the upper part of the figure, the user has to 
% In a conventional setting (diagram's upper half), the user needs to 
separately import three different libraries (\ie \texttt{pandas}, \texttt{scikit-learn}, \texttt{matplotlib}) and use five of their modules. On the other hand, as shown in the lower part of the figure, by using \exekglib, %(diagram's lower half),
% the user only needs a limited number of libraries and modules, 
the user can easily discover and invoke the required functionalities from withing our library only. 
% and thus 
This indeed makes learning easier and faster by skipping reading extensive documentation of various libraries.

By utilizing \rdflib combined with the standard SHACL for describing and validating RDF graphs, and standard PyData (Python for Data) tools, the proposed Python library is capable of:
\begin{enumerate}[topsep=3pt,parsep=0pt,partopsep=0pt,itemsep=0pt,leftmargin=*]
    \item Generating executable ML pipelines as KGs, covering (a) data visualization, (b) data preprocessing and feature engineering, and (c) model training and testing.
          % \begin{itemize}
          %     \item Data visualization
          %     \item Data preprocessing and feature engineering
          %     \item Model training and model testing
          % \end{itemize}
    \item Validating the constructed KGs to guarantee their executability.
    \item Transforming the constructed KGs to Python scripts and executing them.
\end{enumerate}

While rich in methods for the above tasks, the library leverages semantic technologies for convenient extendability by allowing: %in the future regarding the following aspects:
\begin{enumerate}[topsep=3pt,parsep=0pt,partopsep=0pt,itemsep=0pt,leftmargin=*]
    \item Introducing more ML modeling methods.
    \item Performing customized feature engineering and data visualization.
    \item Using existing external packages in custom methods.
\end{enumerate}

\subsection{Underlying KG Schemata}
\label{sec:kg-schemata}

\begin{figure}[t]
% \vspace{-4ex}     
     \centering
     \includegraphics[width=.9\textwidth]{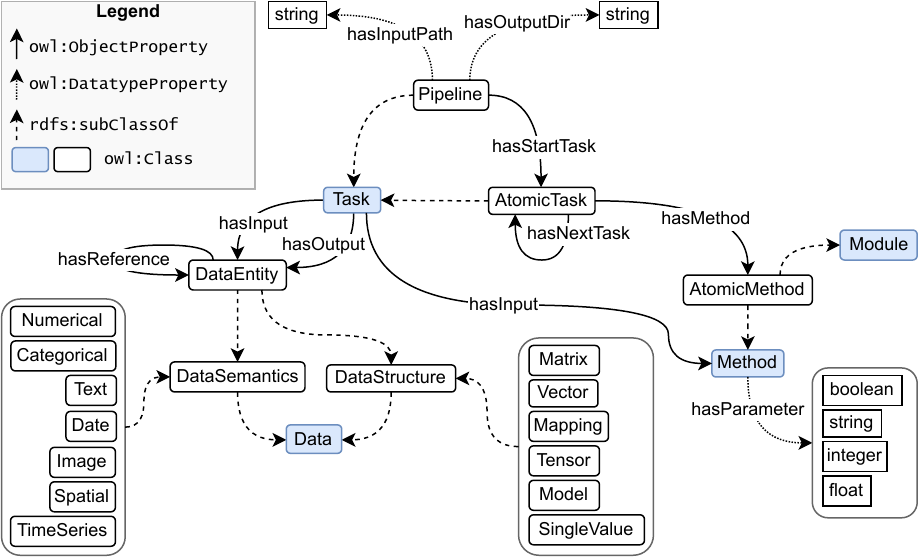}
     % \vspace{-2ex}
     \caption{Data Science (DS) KG schema}
     \label{fig:ds-kg-schema}
      % \vspace{-4ex}
 \end{figure}

\begin{figure*}[t]
% \vspace{-4ex}
\centering
     \includegraphics[width=1\textwidth]{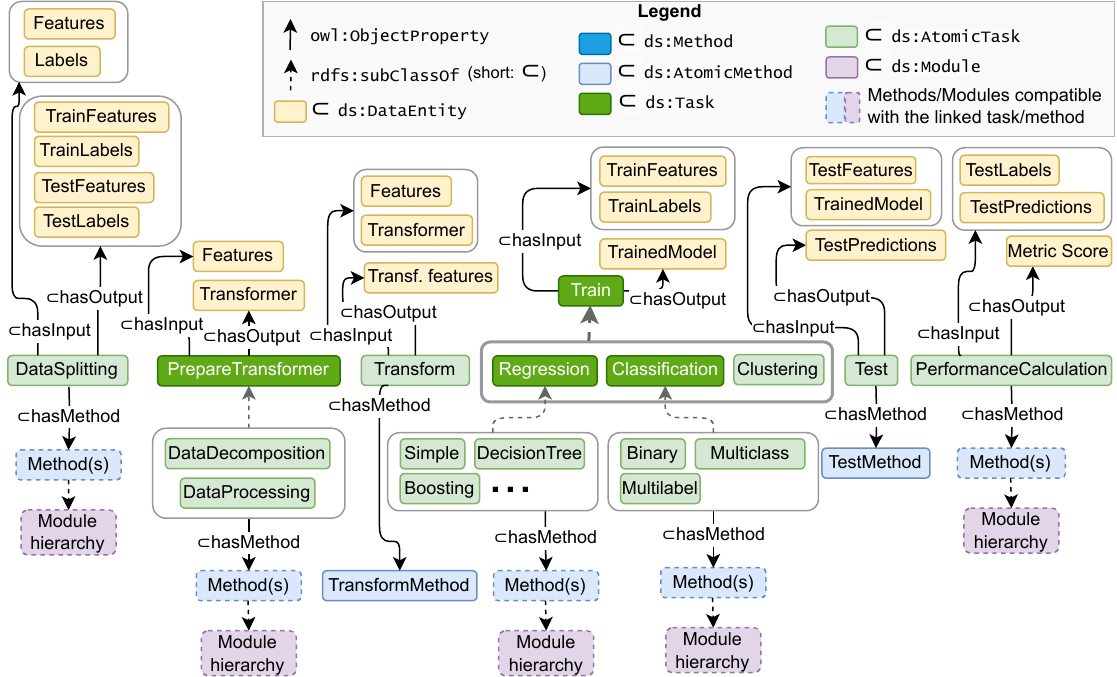}
     % \vspace{-4ex}
     \caption{Machine Learning (ML) KG schema}
     \label{fig:ml-kg-schema}
    % \vspace{-4ex}
 \end{figure*}

\exekglib utilizes one upper-level and three lower-level KG schemata to describe ML, Statistics, and Visualization tasks. The upper-level KG schema is shown in Fig.~\ref{fig:ds-kg-schema}, and of the specific KG schemata for ML, Statistics, and Visualization, the ML KG schema is shown in Fig.~\ref{fig:ml-kg-schema}.
% The other two schemata are similar to Fig.~\ref{fig:ml-kg-schema} and thus omitted for simplicity.
% Readers can read the documentation of \exekglib for details. 
% Schemata for Statistics and Visualization tasks can be found in \exekglib's documentation. 
%The KG schemata are described below.
Following detailed description of the used schemata.

% \begin{figure}[t]
%     \centering
%     \begin{subfigure}[t]{0.47\textwidth}
%          \centering
%          \includegraphics[width=1\textwidth]{content/figures/DSKGSchema.drawio.pdf}
%          \caption{Data Science (DS) KG schema}
%          \label{fig:ds-kg-schema}
%      \end{subfigure}
%      \hfill
%     \begin{subfigure}[t]{0.5\textwidth}
%          \centering
%          \includegraphics[width=1\textwidth]{content/figures/MLKGSchema.drawio.pdf}
%          \caption{Machine Learning (ML) KG schema 
%          % \heiko{Why do we define a specific hasXTrainMethod property -- does that mean there is a specific property for each classifier, such as hasKNNTrainMethod? Why is that necessary?}
%          }
%          \label{fig:ml-kg-schema}
%      \end{subfigure}
%      % \vspace{-2ex}
%      \caption{KG schemata used by \exekglib \heiko{shouldn't the right hand side also repeat data?}}
%      % \vspace{-2ex}
% \end{figure}

% \begin{figure*}[t]
%      \centering
%      \includegraphics[width=1\textwidth]{content/figures/StatsKGSchema.drawio.pdf}
%      \caption{Statistics KG schema}
%      \label{fig:stats-kg-schema}
%         % \vspace{-4ex}
%  \end{figure*}

%  \begin{figure}[t]
%      \centering
%      \includegraphics[width=.45\textwidth]{content/figures/VisuKGSchema.drawio (2).pdf}
%      \caption{Visualization KG schema}
%      \label{fig:visu-kg-schema}
%         % \vspace{-4ex}
%  \end{figure}

% \vspace{-2ex}
\medskip
\noindent \textbf{Data Science (DS) Schema.}
The DS (namespace: \texttt{ds}) KG schema is the upper-level schema to which the rest refer. It represents the general concepts of Data Science. Particularly, it contains the upper level \owlclass entities (in blue): \textit{Data}, \textit{Method}, and \textit{Task}. The \textit{Data} class includes concepts related to data, such as \textit{DataSemantics}, which describes the meaning of data, and \textit{DataStructure}, which specifies the format of data \eg a \textit{Numerical} can have the format \textit{Vector}. The \textit{Method} class includes algorithms and functions (with allowed input, output, and parameters) that operate on data and are indicated by the \textit{AtomicMethod} class. The \textit{Task} class has two sub-classes: \textit{AtomicTask} and \textit{Pipeline}. The former defines the pipeline's tasks, that execute the mentioned functions. The latter is an ordered series of tasks that organize data movement.
% In addition to these classes, there are constraints that apply to certain data, such as the requirement that an \textit{Array} must have a certain number of dimensions.

% \vspace{-2ex}
\medskip
\noindent \textbf{Machine Learning (ML) Schema.}
The ML KG schema (namespace: \texttt{ml}) is an example of a lower-level schema that contains entities of type \owlclass that are sub-classes (\rdfssubclassof) of \textit{AtomicTask} and \textit{AtomicMethod} which are defined in the DS schema. Each lower-level sub-class of \textit{AtomicTask} refers to a task type that can be solved using a specific group of methods (\ie \textit{AtomicMethods}). The boxes with dashed frames can be seen as templates for methods that implement specific algorithms suitable for solving the connected task types. For instance, tasks of type \textit{Classification} can be solved using a method that implements the k-NN algorithm.
% that trains or tests an ML model, calculates the model's performance, or splits the data. 
Task types' organization follows the common ML processes of data splitting, training and testing regression, classification, and clustering models, and calculating the models' performance.

% train or test a specific ML model and the `X' in their name indicates the name of that model.
% Currently, Linear Regression, MLP, and k-NN are supported.
% \heiko{It looks to me that there is a middle layer in the class hierarchy missing. Currently, it seems like there are classes like KNNClassifier which are direct subclasses of TrainTask, but I would rather expect a hierarchy reflecting that there is classification (with flavors like binary/multi-class/multi-label/hierarchical...), regression, ... Also, the X nomenclature raises the impression that there are specific test methods for each classifier (like a test method for KNN, another one for MLP), while I assume it's not like that}
% \antonis{added hierarchy in the figures and in the schema paragraphs. I also replaced the X nomenclature}

% \vspace{-2ex}
\medskip
\noindent \textbf{Statistics and Visualization Schemata.}
The other two KG schemata (namespaces: \texttt{stats} and \texttt{visu} respectively) follow the same structural principles as the ML schema. The difference is the contents of the lower level sub-classes which in this case consist of tools for statistics and data visualization. For the former, the task types are grouped based on the statistical measure to be calculated or the transformation to be applied to the data. Various descriptive statistics can be calculated such as central tendency measures (\eg median), position measures (\eg percentile), and frequency distributions (\eg grouped frequency distributions). 
% Also, one of the statistics task types refers to preparing the data for ML by applying normalization, scaling, etc. 
As for data visualization, the task types are divided depending on whether the user wants to do basic (\eg scatter), statistical (\eg boxplot), or multivariate (\eg heatmap) plotting.
% there are methods that perform normalization, scattering, and trend and mean calculation. For the latter, plotting methods are included to create scatter, line, and bar plots.
% \heiko{This is a journal paper where we have space, so it doesn't hurt to show those two schemas as well}
% \antonis{added}

% In all three lower-level schemata, the user can add any custom sub-class of \dsmethod that will be used by a newly defined sub-class of \dstask. This is also mentioned in section \ref{easy-extension} as a step for conveniently extending the library.

% \antonis{added the below subsection to mention the generated KG schemata. I have created a tool for this. should I further describe the tool in separate section or just keep this subsection?}

% \vspace{-2ex}
\medskip
\noindent \textbf{Semi-automatically Generated Schemata.}
The lower-level KG schemata were partially generated using a semi-automatic process involving popular data science Python libraries such as \texttt{scikit-learn}, \texttt{matplotlib}, and \texttt{numpy}. This process included extracting Python class and method definitions, docstrings, and module hierarchies from the libraries, and then converting them to KG components using our conversion tool and a predefined method-to-task mapping. This approach allows for the generated parts of the KG schemata to be easily updated to accommodate changes in the Python libraries. Additionally, the conversion tool generates SHACL constraints to accompany the converted KG components. This method of populating the KG schemata and SHACL constraints provides a predefined set of methods and tasks for users to choose from, and allows \exekglib to validate ML pipelines.

\subsection{Executable KG Construction}
\begin{figure*}[t]
% \vspace{-4ex}
    \centering
    \includegraphics[width=1\textwidth]{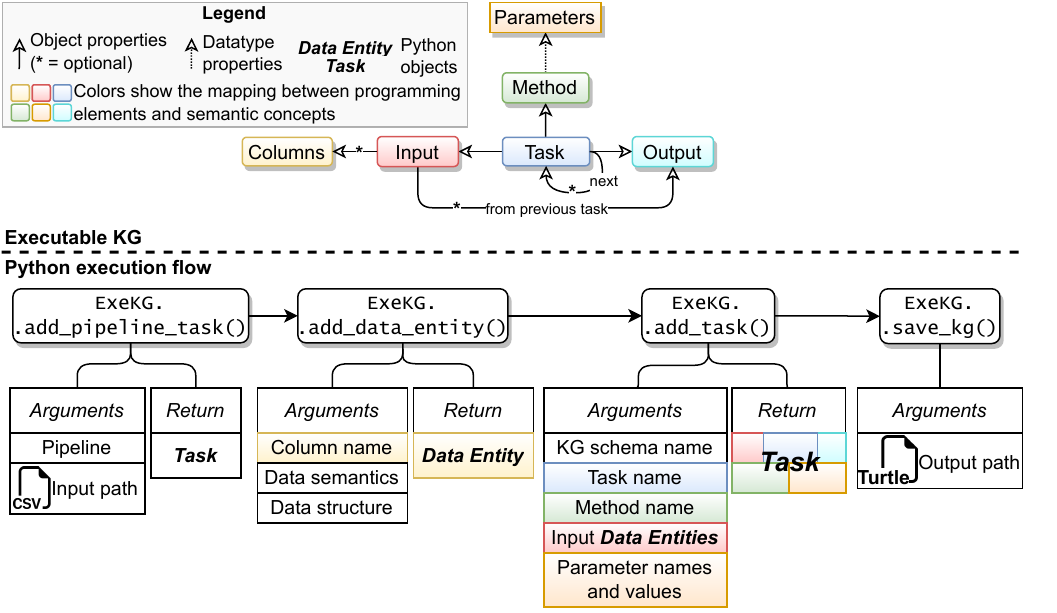}
    % \vspace{-5ex}
    \caption{Executable KG construction. Linking semantic concepts (\eg \textit{Task} entity) with programming elements (\eg \textit{Task} object); Sample code in Fig.~\ref{fig:code-comparison}.}
    \label{fig:kg-construction}
    % \vspace{-4ex}
\end{figure*}

\exekglib supports 
the creation of an ExeKG either programmatically or via the provided \texttt{Typer} CLI. This can be also achieved using the GUI (Section~\ref{sec:gui}).
% \heiko{We have a slight glitch in our argumentation here. We compare to tools which have an intuitive graphic UI, while proposing something with a much steeper learning curve. I think this is not really easy to resolve (unless we build a UI), so we need to be careful how we sell ExeKGLIb using other advantages.}
% \antonis{I added a brief mention to the GUI. I did the same for the subsection: ML pipeline execution}
% In this section, we focus on the programmatic usage of \exekglib %because it is a suitable basis to understand 
% to offer better understanding of 
% how the tool works under the hood. 
For the sake of a more comprehensive understanding of \exekglib's underlying mechanisms, below we focus on the programmatic usage of the library.  
The internal process of creating an ExeKG is illustrated in Fig.~\ref{fig:kg-construction} and is described below. As a prerequisite, a Python object of class \textit{ExeKG} should be instantiated by the user, for \exekglib to create an empty KG and retrieve and parse the KG schemata.

\begin{sloppypar}
At first, the user should provide the path of the input CSV file using the \texttt{ExeKG.add\_pipeline\_task()} Python method. Based on this file, the user can add data columns to the KG using the \texttt{ExeKG.add\_data\_entity()} Python method. Under the hood, \exekglib populates the KG with \dsdataentity \owlindivs representing the target columns. These can be later used as input to the ML pipeline tasks. 
\end{sloppypar}

With the data already defined, the user can specify the operations to perform on the data. This is done by using the \texttt{ExeKG.add\_task()} Python function for each operation. The user should first choose a task to perform, the name of which corresponds to a sub-class of \dstask \owlclass. Then, they should select a method compatible with this task, the name of which corresponds to a sub-class of \dsmethod \owlclass. Afterward, the user should decide which will be the input data entities for this pipeline task. They can be \dsdataentity \owlindivs representing the input CSV columns or the output of previously added ML pipeline tasks. To parametrize the manipulation of the data, the user can specify values for the parameters of the chosen method. These parameters correspond to datatype properties that are linked to the chosen \dsmethod's sub-class. As soon as \texttt{ExeKG.add\_task()} is called, \exekglib adds to the KG the \owlindivs representing the user-specified task (\eg classification) and method (\eg k-NN) and links the current task with the chosen method, input data entities, datatype properties, and the next ML pipeline task. The user does not need to have any knowledge about the structure of \exekglib's KG schemata, since the names of all the needed KG components are presented in a user-friendly way in the tool's documentation.

Finally, with a call to \texttt{ExeKG.save\_kg()}, \exekglib serializes the created KG and saves it on the disk in Turtle.

% As shown in Figure \ref{fig:kg-construction}, the internal process of creating an ExeKG starts with extracting the columns from the input dataset (CSV file). \exekglib populates the KG with \textit{DataEntity} \owlindivs representing the target columns. These \owlindivs are then used as input to the ML pipeline tasks.

% Afterward, \exekglib adds to the KG the entities representing the user-specified task type (\eg classification) and method type (\eg k-NN), which are taken from the provided lower-level KG schemata; and links the current task with the chosen method, input data entities, datatype properties, and the next task. Finally, the created KG is serialized and saved on the disk in Turtle.

% \medskip
% \noindent \textbf{Executable KG validation.}
\subsection{Executable KG Validation}
\label{exe-kg-validation}
% \baifan{I will ask Xun to check this section. He is doing a paper specifically on the topic. Now it has a mixture perspective of semantics and syntax. I think it makes sense to only talk from the semantics  view  (pipeline structure, pipeline attribute) or syntax (object property, range, datatype property and literals), or both and make correspondance table.}
By performing KG validation under the hood, we minimize the chance of user error and give the appropriate feedback to the user during the creation of ExeKGs. \exekglib uses shape constraints to ensure the validity and executability of the constructed KGs. The initial SHACL shape graph was generated using \texttt{sheXer} Python library \cite{fernandez-alvarezAutomaticExtractionShapes2022}, which can automatically extract SHACL shape constraints for RDF graphs. At first, the generated shape graph was slightly modified to reflect the constraints inferred by the KG schemata, \eg from \texttt{rdfs:range} property. With this shape graph as a baseline, special constraints required for the case of \exekglib were manually added. These constraints can be categorized according to the below three aspects of an ExeKG.

% Gad, to have more consistency I will replace the description/item interface with paragraphs, ok?
% \medskip
% \noindent \textbf{Automatic mapping of KG components to programming components for easy extension.}
% like this

% \vspace{-2ex}
\medskip
\noindent \textbf{Pipeline Structure.}
For a valid ML pipeline, a series of \dstask \owlindivs which invoke specific \dsmethod \owlindivs should be connected with each other. Some examples are: (a) Each \dsatomictask \owlindiv must be followed by at most one \dsatomictask \owlindiv via \dshasnexttask, (b) each \dsatomictask \owlindiv must have exactly one compatible \dsatomicmethod \owlindiv connected to it. 
In addition, the series of \dstask \owlindivs of specific \dsatomictask sub-classes should be in a particular order. For instance, before any \plottask \owlindiv, a \canvastask \owlindiv has to be added to the pipeline because it defines the grid layout for the plots.

% \begin{lstlisting}[basicstyle=\footnotesize\ttfamily, caption={Example constraint 1: Each pipeline task must be followed by at most one task.}, label={lst:pipeline-constraint}]
% :AtomicTask a sh:NodeShape ;
%     sh:property [ a sh:PropertyShape ;
%             sh:maxCount 1 ;
%             sh:path ds:hasNextTask ] ;
%     sh:targetClass ds:AtomicTask .
% \end{lstlisting}

% \begin{lstlisting}[basicstyle=\footnotesize\ttfamily, caption={Example constraint 2: Each pipeline task must be linked to a compatible method.}, label={lst:task-method-constraint}, breaklines=true]
% :PerformanceCalculationTaskMethodShape a sh:NodeShape ;
%     sh:targetClass ml:PerformanceCalculation ;
%     sh:property [
%         sh:path ml:hasPerformanceCalculationMethod ;
%         sh:minCount 1 ;
%         sh:maxCount 1 ;
%         sh:or (
%             [ sh:class ml:AccuracyScoreMethod ]
%             [ sh:class ml:F1ScoreMethod ]
%             ...
%             [ sh:class ml:PrecisionScoreMethod ]
%             [ sh:class ml:R2ScoreMethod ]
%         ) ;
%         sh:message "Tasks of type PerformanceCalculation must be connected with exactly one compatible atomic method." ;
%     ] .
% \end{lstlisting}

% \vspace{-2ex}
\medskip
\noindent \textbf{Data Entities.}
The tasks of a pipeline are related to \dsdataentity \owlindivs that represent their inputs and outputs, which are constrained based on the \dstask \owlindivs. In particular, the type of data represented by these \dsdataentity \owlindivs is determined using the sub-classes of \dsdatastructure \owlclass as shown in Fig.~\ref{fig:ds-kg-schema}. This allows for enforcing SHACL constraints such as: The output of each \traintask \owlindiv must be a \dsdataentity \owlindiv, representative of the trained ML model, that is connected to the \dssinglevalue sub-class of \dsdatastructure. Besides, the number of \dsdataentity \owlindivs for some \dstask \owlindivs is also constrained. For instance, each \traintask \owlindiv must have at least two inputs (representing data for features and labels) and only one output. In this case, the third input is optional as it is used only by \trainmethod \owlindivs that correspond to ensemble models or hyperparameter tuners.

% \begin{lstlisting}[basicstyle=\footnotesize\ttfamily, caption={Example constraint 3: Each training task must have a single value data entity as an output.}, label={lst:dataentity-constraint}, breaklines=true]
% :TrainTaskInputOutputShape a sh:NodeShape ;
%     sh:targetClass ml:Train ;
%     # input
%     sh:property [
%         sh:path ml:hasTrainInput ;
%         sh:qualifiedMinCount 1 ;
%         sh:qualifiedMaxCount 1 ;
%         sh:qualifiedValueShape [
%             sh:class ml:DataInTrainX ;
%         ] ;
%         sh:message "Tasks of type Train must be connected with exactly one input of type DataInTrainX." ;
%     ] ;
%     sh:property [
%         sh:path ml:hasTrainInput ;
%         sh:qualifiedMinCount 1 ;
%         sh:qualifiedMaxCount 1 ;
%         sh:qualifiedValueShape [
%             sh:class ml:DataInTrainY ;
%         ] ;
%         sh:message "Tasks of type Train must be connected with exactly one input of type DataInTrainY." ;
%     ] ;
%     sh:property [
%         sh:path ml:hasTrainInput ;
%         sh:qualifiedMaxCount 1 ;
%         sh:qualifiedValueShape [
%             sh:class ml:InputModelAsMethod ;
%         ] ;
%         sh:message "Tasks of type Train must be connected with at most one input of type InputModelAsMethod." ;
%     ] ;
%     # output
%     sh:property [
%         sh:path ml:hasTrainOutput ;
%         sh:minCount 1 ;
%         sh:maxCount 1 ;
%         sh:class ml:DataOutTrainModel ;
%         sh:message "Tasks of type Train must be connected with exactly one output of type DataOutTrainModel." ;
%     ] .
% \end{lstlisting}

% \vspace{-2ex}
\medskip
\noindent \textbf{Pipeline Attribute Values.}
The attribute values are represented as literals in an ExeKG. So, SHACL constraints are used for the type of literal values, based on the standard XML Schema Definition (XSD). As an example, \dsmethod \owlindivs that represent MLP classifiers must have at most one integer literal for `batch size'.
% (Constraint~\ref{lst:attribute-constraint}).

% Naturally, the types of literal values are constrained via SHACL by using the standard XML Schema Definition (XSD). However, for some datatype properties that are attached to \dsmethod \owlindivs, additional constraints are needed. As an example, the value of the datatype property representing the grid layout for the \canvastask \owlindiv must be a string and also have a specific regex pattern: `[0-9] [0-9]' (Constraint~\ref{lst:attribute-constraint}).
% \heiko{It would be good to have a listing with 1-2 exemplary SHACL constraints}
% \antonis{added}

% \begin{lstlisting}[basicstyle=\footnotesize\ttfamily, caption={Example constraint 4: The `batch size' parameter for the MLP classifier must be an integer.}, label={lst:attribute-constraint}, breaklines=true]
% :MLPClassifierMethodParameterShape a sh:NodeShape;
%     sh:targetClass ml:MLPClassifierMethod ;
    
%     sh:property [
%         sh:path ml:hasParamBatchSize ;
%         sh:maxCount 1 ;
%         sh:or (
%             [ sh:datatype xsd:int ]
%         ) ;
%         sh:message "Method MLPClassifierMethod must have at most one compatible value for parameter hasParamBatchSize." ;
%     ] .
% \end{lstlisting}

\subsection{ML Pipeline Execution}
Similar to ExeKG creation, ML pipeline KGs can be executed via code, CLI, or the GUI (Section~\ref{sec:gui}). To execute a KG, \exekglib parses the KG using the above KG schemata (Section~\ref{sec:kg-schemata}). After that, the pipeline's \dstask \owlindivs are sequentially traversed using the object property \dshasnexttask. Based on the IRI of the next \dstask \owlindiv, the \owlindiv's properties are retrieved and mapped dynamically to a Python object. Such mapping allows for extending the library without modifying the KG execution code. Finally, for each \dstask \owlindiv, the Python implementation that corresponds to the linked \dsmethod \owlindiv is invoked.

The internal process that \exekglib uses for executing a pipeline can be divided into the below steps:
\begin{enumerate}[topsep=3pt,parsep=0pt,partopsep=0pt,itemsep=0pt,leftmargin=*]
% \medskip
% \noindent \textbf{KG and Dataset Loading.}
        \item \textbf{KG and Dataset Loading}:
        The pipeline is in the form of an ExeKG, so the KG is parsed using the \rdflib package to build an \textit{ExeKG} Python object. The \dspipeline \owlindiv is retrieved from the KG and after parsing it, \exekglib loads the CSV dataset from the path indicated by the \dshasinputdatapath data type property.
% \medskip
% \noindent \textbf{Parsing of \dstask \owlindiv.}
        \item \textbf{Parsing of \dstask \owlindiv}:
        \exekglib starts traversing the pipeline using the object property \dshasnexttask. For each \dstask \owlindiv, the attached \dsmethod \owlindiv, input and output \dsdataentity \owlindivs, and \dsmethod's datatype properties are extracted and stored in a Python object for convenient access.
% \medskip
% \noindent \textbf{Execution of \dstask \owlindiv.}
        \item \textbf{Execution of \dstask \owlindiv}:
        The Python object created while parsing the \dstask \owlindiv and its connected components, contains an abstract Python method (\texttt{run\_method()}) that implements the functionality indicated by the names of the \dstask and \dsmethod \owlindivs. This method is called internally by \exekglib and its arguments are (1) the parsed input \dsdataentity \owlindivs that have been translated to outputs of the previous pipeline's tasks or to columns of the input CSV dataset, and (2) the parsed \dsmethod datatype properties which have been translated to Python primitive constants.
\end{enumerate}

\medskip
% \noindent \textbf{Automatic Mapping of KG Components to Programming Components for Easy Extension.}
\noindent \textbf{Automatic KG-to-Code Mapping for Easy Extension.}
During the traversal of an ExeKG, each \dstask \owlindiv is automatically mapped to a Python class, and a Python object is created. 
% For this to happen, the name of the child \textit{Task} class must be the concatenation of the KG task name with the name of its KG method. For instance, a task with name \textit{PlotTask} that has a method with name \textit{LinePlotMethod} in the KG, will be instantiated as an object of the Python class with name \textit{PlotTaskLinePlotMethod}.
As for the properties of the \dstask \owlindiv, they are also automatically mapped to the corresponding class fields. Most properties' values (\ie IRIs or data [\eg of type string]) are converted by Python under the hood when they are assigned using the \texttt{setattr()} built-in function. However, the IRIs of \dsdataentity \owlindivs require a special treatment due to the \textit{Referenced IRI} field described in Paragraph \hyperref[par:data-entity-class]{\textit{Data Entity} Class}. As a result, the user can extend the library to accommodate new tasks with custom methods without having to modify the code used for KG construction or KG execution.

\subsection{Object-oriented Programming Architecture}

\begin{figure*}[t]
    % \vspace{-4ex}
    \centering
    \includegraphics[width=0.9\textwidth]{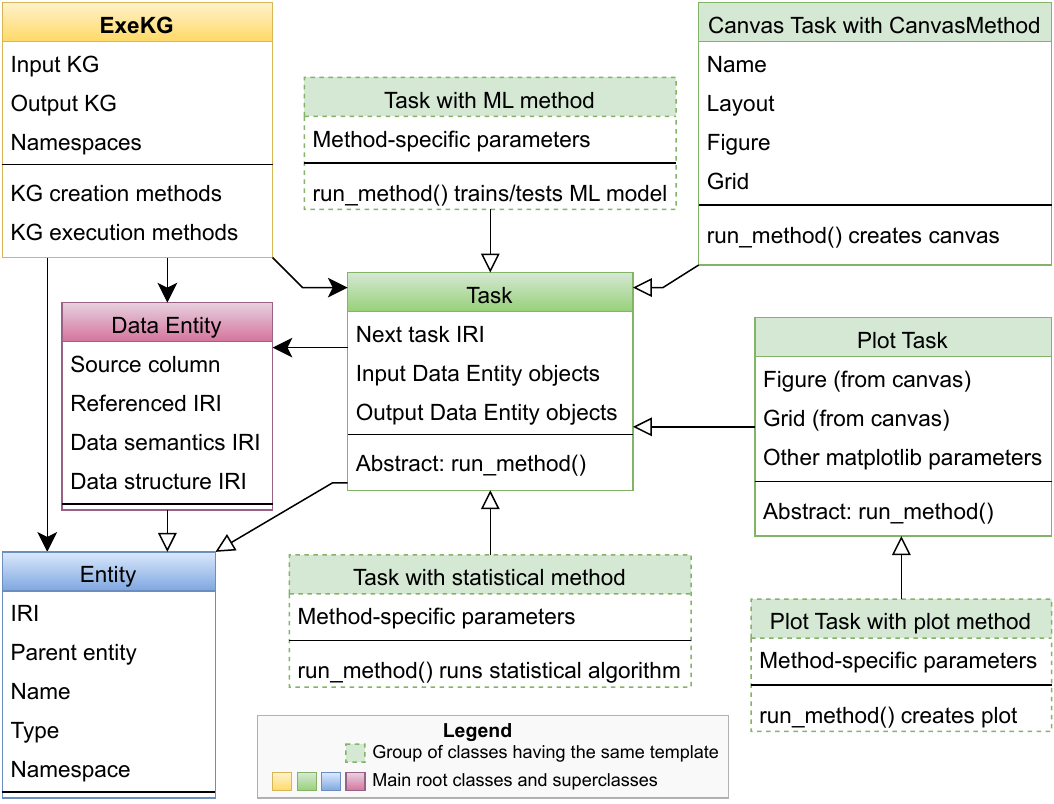}
    % \vspace{-3ex}
    \caption{High-level architecture of the class (object-oriented programming) hierarchy of \exekglib expressed with a UML Class Diagram}
    \label{fig:uml-diagram}
    % \vspace{-6ex}
\end{figure*}

To achieve temporary storage of information found in the KGs, a custom Python class hierarchy was built (Fig.~\ref{fig:uml-diagram}). The classes used are abstractions of KG entities and contain fields that map to KG properties. \exekglib instantiates objects of these classes to save and access parsed information about \owlindivs and \owlclasses of the KGs and KG schemata. The main classes are described below.

% \vspace{-2ex}
% \vspace{-1ex}
\medskip
\noindent \textbf{\textit{ExeKG} Class.}
The main Python class is \textit{ExeKG} and has various fields including custom \textit{Namespaces} and two \rdflib Graph objects: \textit{Input KG} and \textit{Output KG}. In the case of KG construction, the \textit{Input KG} contains all KG schemata used as a basis to create the ML pipeline, and the \textit{Output KG} is the constructed ExeKG representing the pipeline. \textit{KG creation methods} are a set of methods that add components (\eg \owlindivs) to the executable \textit{Output KG}. High-level \textit{KG creation methods} (\eg \texttt{ExeKG.add\_task()}) are for users to conveniently build KGs programmatically. During the process of ML pipeline execution, the \textit{Input KG} is the ExeKG and the \textit{Output KG} is not used. For running the created ML pipeline, the \textit{KG execution methods} are used under the hood to traverse the KG, convert each task with its specified method and properties to Python code, and execute it. For this reason, the \textit{Task} class has been created together with child classes that represent the different task-method combinations.

% \vspace{-1ex}
\medskip
\noindent \textbf{\textit{Task} Class.}
Except for the inherited components of the \textit{Entity} class, the \textit{Task} class also contains a \textit{Next Task IRI} field pointing to the next \textit{Task} of the pipeline, the \textit{Input Data Entity objects}, and the \textit{Output Data Entity objects}. The latter two are Python dictionaries that contain the correspondence between a task's input names and objects of the \textit{Data Entity} class. These objects represent relevant entities of the KG. In the case of \textit{Input Data Entity objects}, they can refer to the pipeline's initial input data or output data from previous tasks of the pipeline. The \textit{Task} class has also a \texttt{run\_method()} Python method which is abstract and is implemented by child classes to execute a specific algorithm. The child classes are \textit{Tasks} that are associated with different ML-related methods. There are \textit{Tasks} implementing algorithms for ML (\eg Linear Regression model training and testing), Statistics (\eg normalization), and Data Visualization (\eg line plot). For the last class group, there is \textit{Plot Task} as a common parent that stores information about the canvas that is used by the child \textit{Plot Task} classes. The set of canvas parameters (\textit{Name}, \textit{Layout}, \textit{Figure}, \textit{Grid}) is initialized by the \texttt{run\_method()} of the \textit{Canvas Task with Canvas Method} class. 
% This task is mandatory to be executed before any \textit{Plot Task} runs because the \textit{Plot Task} needs the \textit{Figure} and \textit{Grid} canvas parameters to define the plot's drawing space. 
% The \textit{Plot Task} classes provide built-in support for using matplotlib.

% \vspace{-2ex}
\medskip
\noindent \textbf{\textit{Data Entity} Class.}
\label{par:data-entity-class}
The \textit{Data Entity} class refers to the data that are used as input or output for the pipeline's tasks. Its \textit{Source column} field is the name of a column from the pipeline's input CSV file and its \textit{Referenced IRI} field acts as a link to an output data entity of a previous task in the pipeline. In case this class is used to represent input, only one of the two mentioned fields has a value. Furthermore, the \textit{Data Entity} class includes the fields \textit{Data semantics IRI} and \textit{Data structure IRI} that associate the data with their structure (\eg vector) and semantics (\eg time series).

% \vspace{-2ex}
\medskip
\noindent \textbf{\textit{Entity} Class.}
The \textit{Entity} class is an abstraction of a KG entity and is the parent class of \textit{Data Entity} and \textit{Task} classes. It consists of basic information about a KG entity: its \textit{IRI}, \textit{Name} (the part after `\#' in the \textit{IRI}), \textit{Type} (\textit{Parent entity}'s \textit{Name}), \textit{Parent Entity} and \textit{Namespace} (the part before `\#' in the \textit{IRI}). It is directly used by \textit{ExeKG}'s \textit{KG creation methods} and \textit{KG execution methods} as a means of temporary storage of information for the parsed KG entities.

\subsection{Graphical User Interface}
\label{sec:gui}
Leveraging \exekglib's LOD framework, the GUI (Fig.~\ref{fig:gui-elements}) facilitates the creation of reusable and interoperable ML pipelines. It employs KGs and data science ontologies for a structured and transparent representation of ML workflows, explicitly encoding relationships between tasks, methods, and datasets to ensure reusability and executability~\cite{klironomos2023exekglib}. The integration of intuitive visual design, an intelligent AI assistant, and ontology-driven structuring significantly lowers technical barriers, making advanced ML accessible to a broader audience.

\medskip
\noindent \textbf{Interactive Graphical Features.}
The GUI (Fig.~\ref{fig:gui-elements}) enables interactive ML workflow creation and execution. Users visually construct pipelines by dragging task and feature nodes from a sidebar (which includes a search function) onto a canvas and connecting them with directed edges to define flow and dependencies. The graph-based canvas allows easy modification of the pipeline structure, and task behavior can be adapted through associated method and parameter nodes. A `Run Pipeline' button initiates the transformation of the visual graph into an ExeKG object, which is sent to the backend for execution. Results are then displayed in the GUI.

\begin{figure*}[t]
    % \vspace{-4ex}    
    \centering
    \includegraphics[width=1\textwidth]{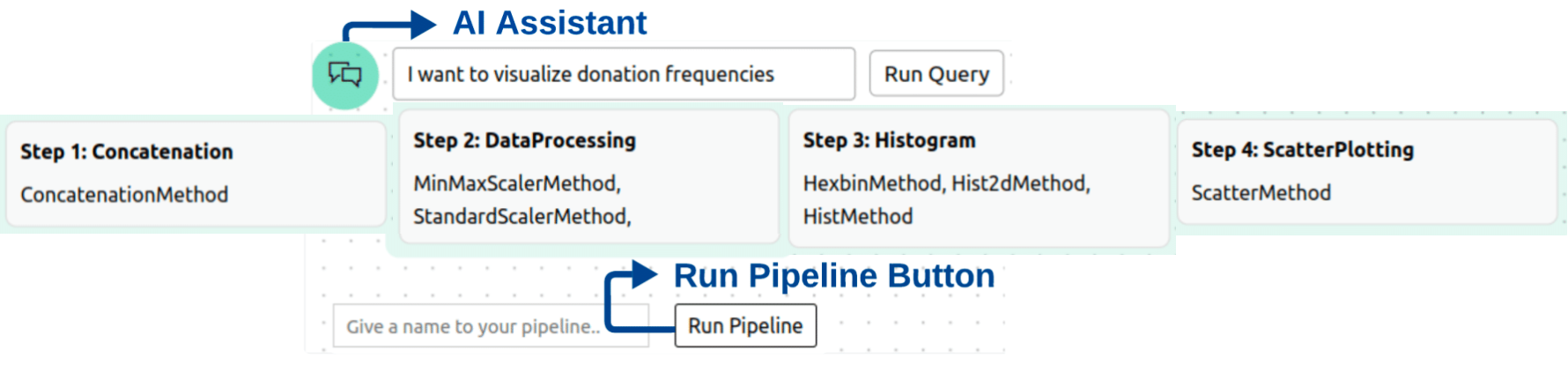}
    % \vspace{-6ex}
    \caption{Prompt-based LLM-Powered Recommendations}
    \label{fig:assistant}
    % \vspace{-6ex}    
\end{figure*}

\medskip
\noindent \textbf{LLM-powered recommendation engine.}
An LLM-powered AI assistant (Fig.~\ref{fig:assistant}) helps users, especially non-experts, construct ML pipelines by interpreting natural language queries. Integrated via a chatbox, it provides tailored pipeline recommendations (tasks and methods) based on user input and dataset metadata, guiding users to build workflows directly on the canvas.
% A prompt-based approach was adopted for its superior accuracy in generating these suggestions.
% \baifan{sounds very comprehensive, but quite succinct in details. where to find more information? or maybe a small screenshot of the chatbot in fig.7?}
% \antonis{changed the figure}

% \antonis{commented out 'Evaluation of LLM-based Recommendations' paragraph}
% \medskip
% \noindent \textbf{Evaluation of LLM-based Recommendations.}
% Evaluation of the LLM recommendations against OpenML benchmarks confirmed the chosen prompt-based approach's superior accuracy. For instance, it achieved a Hit Rate of up to 0.60 in suggesting relevant methods found in benchmark pipelines, demonstrating its promise in providing effective guidance. Task overlap, measured by the Jaccard Coefficient, was 0.35; this was attributed to the limited variety of ML tasks in the benchmark dataset compared to the broader range supported by \exekglib. Overall, the LLM integration, particularly with the prompt-based strategy, shows strong potential for providing contextually relevant guidance in ML pipeline construction.
% \input{content/gui.tex}
\section{Impact and Use Cases}
\label{sec:evaluation}

This section explores the tangible impact of \exekglib in industrial settings, detailing its successful implementation at Bosch across various manufacturing applications and its significant contributions to ongoing European research projects. Furthermore, it delves into specific industrial use cases at Bosch, particularly within the domain of welding quality monitoring, to illustrate the practical application and benefits of \exekglib for engineers and ML experts.

\subsection{Impact}
%\evgeny{Here needs Evgeny's input}
%\baifan{I think this section should be extended with enumerating adoption and dissemination of the tool at Bosch and in EU projects}
\exekglib has been successfully evaluated at Bosch under the umbrella of new generation manufacturing monitoring solutions based on neuro-symbolic methods. In particular, \exekglib is the backbone of our semantic machine learning solution, SemML~\cite{DBLP:conf/cikm/ZhouZZSSKK22,zhouSemMLFacilitatingDevelopment2021} that spans over three sub-projects: the resistance spot welding quality monitoring, process optimization for hot-staking, and plastic data analytics (detailed in Sect.~\ref{sec:use-cases}).

Furthermore, \exekglib is an important part of two EU projects \textit{OntoCommons}\footnotehref{https://ontocommons.eu/}
and 
\textit{Graph Massivizer}\footnotehref{https://graph-massivizer.eu/}.
% \heiko{add footnotes with web sites of the two projects}
\textit{OntoCommons} aims to standardize semantic artifacts (including ontologies, KGs, documents, etc.) and find the best practice for creating and maintaining the semantic artifacts. \exekglib is an effort to standardize ML practice and documentation in KGs, improve the transparency and usability of ML solutions for learners of ML that are non-ML experts, such as welding experts, engineers, etc., and facilitate communications between ML experts and non-ML experts.
For instance, \exekglib has shown its great potential for creating ExeKGs that run ML pipelines for  welding quality monitoring and optimization for resistance spot welding and hot staking~\cite{DBLP:journals/jim/ZhouPRKM22,DBLP:conf/cikm/ZhengZZSK22a}  as well as data visualization, and statistic analysis~\cite{zhengExecutableKnowledgeGraphs2022a}.

\textit{Graph Massivizer} aims at creating a platform for information processing and reasoning using large-scale graphs. 
% The platform includes five open-source tools and FAIR graph datasets for the sustainable processing of extreme data. These tools will handle: (1) graph creation, (2) graph enrichment, query, and analytics, (3) graph workload modeling and optimization, (4) sustainability analysis, and (5) serverless processing of basic graph operations (BGO). 
% GAD: starting from here this is pretty good!
In this EU project, four real-world use cases are selected as the basis for developing and verifying the platform. For building the platform's parts where ML is useful, various ML pipelines with different combinations of steps should be created, modified, and tested. \exekglib provides a means of easily creating and storing the ML pipelines in the form of KGs so that they can be easily reused and understood (through visualization). That way, before the deployment of the platform, the project's stakeholders from different disciplines can conveniently compare ML pipelines so that they determine which one is the most suitable for each use case.

Based on the scenarios and use cases, we summarize the impact and benefits of \exekglib: (1) It has a wide range of potential users and scenarios both in the industry and academia; (2) Its open-source nature allows people of different disciplines to use and better understand ML, compared to other similar ML tools that are not open source; (3) The library’s semantic aspect improves the transparency and the tool can bridge the knowledge gap between domain experts and ML specialists; (4) By lowering the barrier of using ML and democratize ML to a wider public, this library can contribute to increasing society’s trust in AI and appreciation towards semantic technologies. % the last phrase is used to refer to the criteria "Will the resource have an impact, especially in supporting the adoption of Semantic Web technologies?"

\subsection{Industrial Use Cases at Bosch}
\label{sec:use-cases}

\begin{figure*}[t]
    \centering
% \vspace{-4ex}    
    \includegraphics[width=1\textwidth]{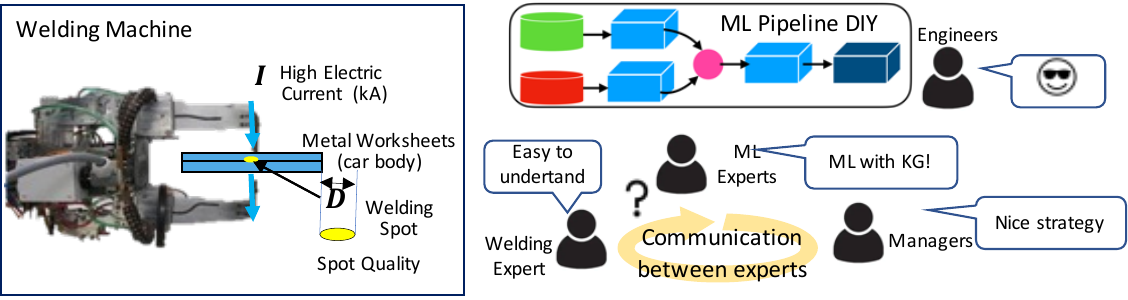}
    % \vspace{-5ex}
    \caption{Use Cases: (1) Engineers create and modify ML pipelines; 
    % \heiko{The text talks about the Q value a lot. Would it be possible to visualize it in this figure?}
    % \antonis{the Q-value is derived by multiple other values, so it cannot be visualized. But I added a reference to the welding spot diameter (D) in the use case 1's section, to make a connection with the Q-Value.}
    (2) Domain experts use semantically annotated data for ML.
    }
    \label{fig:usecase}
    % \vspace{-6ex}
\end{figure*}

The use cases presented in this section, along with their detailed evaluation, have been published in our prior work~\cite{zhouSemMLFacilitatingDevelopment2021,zhengExecutableKnowledgeGraphs2022a}. We include a concise summary of these findings to ensure the paper is self-contained and to provide essential context for the new contributions presented herein.
\exekglib has been used and verified by Bosch in several projects mainly for monitoring welding quality. In the automotive industry, ensuring the quality of welding is critical to the overall performance and safety of vehicles. Bosch has a number of interdisciplinary projects of ML for quality monitoring. We introduce two cases here (Fig.~\ref{fig:usecase}): (1) Bosch engineers learned basic knowledge of ML and want to use ML for quality monitoring; (2) ML experts try to automate ML workflow and need to explain ML to the experts of other disciplines, such as welding engineers, material scientists, managers, etc.

\medskip
\noindent \textbf{Use Case 1: Engineers Create and Modify ML Pipelines.}
% \textbf{Use Case 1: Engineers create and modify ML pipelines.}
As a standalone, this tool has been tested by Bosch in real-world scenarios to predict the quality of resistance spot welding~\cite{zhengExecutableKnowledgeGraphs2022a}. Resistance spot welding is a process in which two metal sheets are joined together by the heat generated from the resistance to electrical current flowing through the sheets. A quality indicator called Q-Value is used to quantify the welding quality, instead of using multiple other quality indicators such as the welding spot diameter (D). It was created empirically by Bosch Rexroth using their extensive engineering knowledge and experience. A Q-Value value of 1 is ideal. If the Q-Value is higher or lower than 1, it can indicate problems with the welding process, such as too much energy being used or a lack of quality. To maintain high quality in welding, it is important to use data from past operations and known features of the upcoming operation to predict the Q-Value and take preventive measures if necessary to ensure that the Q-Value stays as close to 1 as possible.

% The ideal Q-Value is 1, which denotes flawless quality. When the Q-Value is greater than 1, it signifies too much energy is being put into the welding process, and when it is lower than 1, it frequently indicates a lack of quality. Typically, Q-Values are calculated using data gathered from production lines. The goal of quality control throughout manufacturing is to keep the Q-Value for every welding spot as near to 1. In the use case, we would like to achieve this by learning Q-Value estimations prior to the welding itself and then taking preventive actions (e.g. altering welding machine parameters, changing welding caps, etc) if the anticipated value is too low. Single features and time series corresponding to prior welding operations and known features of the upcoming welding operation are the data utilized as input for evaluating the quality of the welding operation.

% A user study was conducted with 28 experts in various fields, including ML, welding, and sensor engineering. The study included a series of tasks related to visual, statistical, and ML analytics, which the participants completed both with and without the use of the developed system. The participants were divided into two groups and completed the tasks in a specific order to control for bias. In addition to completing the tasks, the participants also answered questionnaires about their subjective evaluations. The study also included tasks designed to test the modularized reuse of the ExeKGs.\looseness=-1

A user study was conducted with 28 experts in various fields, including ML, welding, and sensor engineering. The user study included a series of tasks (related to visualization, statistics, and ML analytics) that were completed both with and without the use of the system being tested. ML experts explained the tasks to non-ML experts, who then completed the tasks using technical language or by creating, modifying, or merging knowledge graphs through a GUI. The study measured two metrics: the percentage of tasks that were completed and their completion time. The study also recorded the correctness of the answers to single selection questions and compared the actions taken during the tasks with ground truth to measure correctness. % Some participants were unable to complete all of the tasks.

% In the case of “without our system” (T1, 6, 12), the experts communicate with technical language, and the non-machine learning experts will need to perform the tasks. Due to time limit, it is infeasible to do coding during the user study. The non-machine learning experts will answer whether they can finish the tasks with their programming an machine learning knowledge, and estimate the needed time for that. Thus, we will have two metrics: complete percentage and time. In addition, we compare the answers of SSQs with the correct answers and record the correctness (T2, 7, 13). In case of “with our system” (T3, 5, 8, 10, 11, 14, 16–18), the experts communicate using our system and the non-machine learning experts will need to perform the tasks. They do so by creating, modifying or merging knowledge graphs via a GUI.

The results of the user study show that most participants had a high percentage of tasks completed and a high level of correctness when using the proposed system. 
% They also did not need much time to complete the tasks on average, and their understanding of the ML activities improved when using the system for communication.
The study included, among others, a dimension related to communication (called Transparency) and a dimension regarding reusability. The questions and scores for both dimensions are shown in Table~\ref{tab:user-study-results}. Based on the participants' answers, the resulting scores were above 4 for each dimension.
% ML experts were asked if they were confident in helping non-experts develop ML approaches based on the system. Non-ML experts were asked if they found it easy to get a basic understanding of ML approaches based on the system. Also, a common question between both groups was about feeling that the system hampers the communication on ML approaches.
% Based on the people's answers, the resulting score for the Transparency dimension was 4.28. The scores ranged from 1 (disagree), 2 (fairly disagree), 3 (neutral), 4 (fairly agree), to 5 (agree).
This indicates that the use of our system enhanced the communication between practitioners in different fields. In addition, these results show that both ML experts and non-ML experts consider our system's assets reusable.
The use of the system also resulted in a decrease in time needed to complete tasks and an increase in the percentage of tasks completed, as well as making tasks that were previously not possible for non-ML experts now doable.
% The participants also reported that the system improved transparency and had good usability, and they were satisfied with the coverage of tasks and the reusability and modularity of ML pipelines.

% \vspace{-3ex}
\medskip
\noindent \textbf{Use Case 2: Domain Experts Use Semantically Annotated Data for ML.}
% \noindent \textbf{Use Case 2: Semantically enhancing ML workflows for condition monitoring.}
% \noindent \textbf{Use Case 2: Communication between experts.}
The proposed software was used in a larger system~\cite{zhouSemMLFacilitatingDevelopment2021,zhouSemFEFacilitatingML2020,zhouPredictingQualityAutomated2020} aiming to automate the ML workflow creation and execution. Specifically in~\cite{zhouSemMLFacilitatingDevelopment2021}, a prototype of the \textit{SemML} solution was deployed on data from Bosch for automatic welding.
% The data used for the prototype deployment included information about 13,952 welding operations, comprising inputs of 235 single features and 20 time series.
The goal of this deployment was to conduct experiments involving 14 Bosch experts, including Data Scientists, Measurement Experts, and domain experts. The performance of \textit{SemML} was evaluated on two tasks related to welding quality: estimating the welding spot diameter and predicting the quality of future welding operations based on the quality of previous ones.%\looseness=-1

The data was collected and prepared from two representative welding machines, which perform two and four welding programs respectively. The data includes information about 1,998 and 3,996 welding operations and consists of two levels of features: data on the welding time level, including 4 process curves (time series data) measured per millisecond, and data on the welding operation level, including 188 single features (single feature data). The data includes 2.74 million records and 44.61 million items.
Four ML pipelines were developed and stored in a catalog for users to choose from. These pipelines were designed in a general manner and evaluated using a large dataset collected from resistance spot welding plants.
The pipelines consist of two feature engineering strategies (base and advanced) combined with two ML models (linear regression and LSTM). Base-LR, Base-LSTM, Advanced-LR, and Advanced-LSTM can be used to refer to the four ML pipelines.

Users annotated the raw data with domain terms through a GUI before viewing the available ML pipelines and choosing the one they believe is best suited to handle the ML task. The ML model of the selected ML pipeline was then trained and tested. Finally, the results were visualized using plots.
The best model for each welding machine was determined to be Advanced-LR for Welding Machine 1 and Advanced-LSTM for Welding Machine 2. The Mean Absolute Percentage Error (MAPE) of these models was 1.61\% and 1.94\%, respectively. The difference in performance between the two machines may be due to the complexity of the data.

Finally, a user satisfaction survey for our system was conducted with Data Scientists and experts in the field of welding processes.
Regarding the `Communication easiness' dimension of the user study, participants were asked two questions regarding their perceptions of the resulting ontology/mapping (see Table~\ref{tab:user-study-results}). 
Each question was asked to each group of participants: Data Scientists and domain experts. 
% The first question assessed the ease of understanding of the semantically enhanced workflow for the other group. The second question evaluated the effectiveness of the ontology/mapping as a common base for discussion with the other group.
The answers resulted in a score of 4.70 for this dimension. In other words, the users believe that their work outcomes will be easily comprehensible to other experts. This supports our belief that semantics can provide a robust foundation for communication.
The overall results of the survey showed that users generally had a positive impression of the system and found it easy to use and understand.

\begin{table*}[t]
    % \vspace{-4ex}   
    \centering
    \setlength{\tabcolsep}{2.2pt}
    \caption{Results from prior user studies for each use case (C). Scores ranged from 1 (disagree), 2 (fairly disagree), 3 (neutral), 4 (fairly agree), to 5 (agree). The score was inverted for negation sentences. 
    % \baifan{is it reusing the results from the ExeKG paper? maybe standard deviation is also good?}
    % \antonis{results are from published papers: \cite{zhengExecutableKnowledgeGraphs2022a} and \cite{zhouSemMLFacilitatingDevelopment2021}.
    % Note: I added std and rephrased the questions to not be identical with our published papers.}
    }
    \label{tab:user-study-results}
    \begin{tabular}{ccm{6.8cm}p{2cm}c} 
        \hline
        \textbf{C} & \textbf{Q} & \textbf{Question} & \textbf{Dimension} & \textbf{Score $\pm$ Std} \\ \hline
        % UC 1
        \multirow{11}{*}{1} % UC '1' spans 11 rows (Q1 parts, Q2, Q3 parts, Q4 and their respective \cline and empty line for spacing)
        & \multirow{3}{*}{Q1} & (For ML experts) Using the system, I can confidently help non-experts develop ML approaches. & \multirow{3}{=}{Transparency} & \multirow{3}{*}{$4.28 \pm 0.47$} \\ % Dimension & Score span 3 rows (Q1 parts & Q2)
        \cline{3-3}
        & & (For non-ML experts) The system made it easy to grasp basic ML concepts. & & \\
        \cline{2-3} % Q2 is a separate question under Q1's dimension visually here due to \multirow on Dimension/Score
        & Q2 & The system hindered communication about ML approaches. & & \\
        \cline{2-5} % This line will go across all columns from Q to Score
        
        & \multirow{3}{*}{Q3} & (For ML experts) ML pipelines from this system have limited reusability across applications. & \multirow{5}{=}{Reusability} & \multirow{5}{*}{$4.87 \pm 0.36$} \\ % Dimension & Score span next 4 rows (Q3 parts & Q4)
        \cline{3-3}
        & & (For non-ML experts) I would be hesitant to reuse a developed pipeline for a new task. & & \\
        \cline{2-3} % Q4 is a separate question under Q3's dimension visually here
        & Q4 & The system reduces the time needed to reuse developed pipelines. & & \\
        \hline
        
        % UC 2
        \multirow{4}{*}{2} % UC '2' spans 4 rows (Q5, Q6 and their \cline)
        & Q5 & The resulting ontology/mapping is easily understandable by the other group (\ie Data scientists/Domain experts). & \multirow{4}{=}{Communication easiness} & \multirow{4}{*}{$4.70 \pm 0.50$} \\ % Dimension & Score span 2 questions, 4 rows due to the lines
        \cline{2-3}
        & Q6 & The ontology/mapping offers a good common ground for discussion with the other group (\ie Data scientists/Domain experts). & & \\
        \hline
    \end{tabular}
    % \vspace{-4ex}
\end{table*}
\section{Conclusion and Future Work}
\label{sec:conclusion}

\medskip
\noindent \textbf{Conclusion.}
This paper introduces \exekglib, a Python library that allows people with
% coding skills and 
minimal ML expertise to use ML. \exekglib relies on KG construction that complies with KG schemata to improve transparency and to provide a formalized description of ML scripts. 
% Compared to similar tools, \exekglib has advantages 
We elaborated on the library, including KG schemata, the software architecture, the modules of KG construction and ML pipeline execution, and the GUI.
We then demonstrated the practical impact of \exekglib within EU projects and industrial use cases at Bosch.
\exekglib is open source, aiming at lowering the barrier of using ML and democratizing ML to a wider public.
% \medskip \noindent \textbf{Outlook.}
The current scope of \exekglib has more focus on classic ML methods and tasks, such as exploratory data analysis (data visualization), statistic analysis, feature engineering, and classic ML modeling. This is because the users of \exekglib (domain experts etc.) will more likely start with classic ML methods, and \exekglib has been tested in an industrial environment that has the same focus.

\medskip
\noindent \textbf{Future Work.} 
We aim to extend \exekglib by following the below plan:
\begin{enumerate}[topsep=3pt,parsep=0pt,partopsep=0pt,itemsep=0pt,leftmargin=*]
    \item Publishing a GUI (planned for Q4 2025) like the one used internally by Bosch during evaluation so that we improve the usability of \exekglib and broaden the target user groups.
    % \heiko{Please show an example of the GUI, even if it's not public yet. It would make the paper so much stronger.} 
    % \antonis{added dedicated section}
    \item Supporting a wider range of feature engineering and classic ML methods to cover even more needs of our user base.
    \item Supporting more sophisticated neural networks with higher customizability for advanced users.
    \item Integrating \exekglib with a graph-based database to allow for easier management of the produced ExeKGs, quick visualization, and more convenient reuse.
\end{enumerate}

% The support for more sophisticated neural networks require higher customizability and more rigorous check of the correctness of ML pipelines, and we plan to extend \exekglib in this aspect.  We also plan to build a system by integrating \exekglib with a graph-based database. This will allow for easier management of the produced ExeKGs, quick visualization, and more convenient reuse.

% supporting feature engineering and classic ML -> supporting simple deep learning -> supporting advanced deep learning wherever possible.
% The board and team for managing pull requests from Bosch or two EU projects is currently under discussion.

% Also, we plan to publish a GUI like the one used internally by Bosch during evaluation. We recognize that a GUI will improve the usability of ExeKGLib and broaden the target user groups.
% We also plan to verify the library with more use cases at Bosch. The quality monitoring of several other welding such as hot staking, ultrasonic welding, laser welding will also benefit from \exekglib.

% {\small \noindent \subsubsection*{Acknowledgements}:
% The work was partially supported by EU projects Dome 4.0 (GA 953163), OntoCommons (GA 958371), DataCloud (GA 101016835), Graph Massiviser (GA 101093202) and EnRichMyData (GA 101093202) and the SIRIUS Centre, Norwegian Research Council project number 237898.}

% \begin{credits}
\subsubsection{\ackname} The work was partially supported by EU projects GraphMassivizer (GA 101093202), SMARTY (GA 101140087), and enRichMyData (GA 101070284).
% \subsubsection{\discintname}
% It is now necessary to declare any competing interests or to specifically
% state that the authors have no competing interests. Please place the
% statement with a bold run-in heading in small font size beneath the
% (optional) acknowledgments\footnote{If EquinOCS, our proceedings submission
% system, is used, then the disclaimer can be provided directly in the system.},
% for example: The authors have no competing interests to declare that are
% relevant to the content of this article. Or: Author A has received research
% grants from Company W. Author B has received a speaker honorarium from
% Company X and owns stock in Company Y. Author C is a member of committee Z.
% \end{credits}

\paragraph{Resource Availability Statement:} Source code for \exekglib is available from Github\footnotehref{https://github.com/boschresearch/ExeKGLib}.

% ---- Bibliography ----
%
% BibTeX users should specify bibliography style 'splncs04'.
% References will then be sorted and formatted in the correct style.
%
\bibliographystyle{splncs04}
\bibliography{references}

\end{document}